\documentclass[letterpaper, 10 pt, conference]{ieeeconf}  

\IEEEoverridecommandlockouts                              

\title{\LARGE \bf InfraGPT Smart Infrastructure: An End-to-End VLM-Based Framework for Detecting and Managing Urban Defects
}

\author{
\begin{tabular}{c c}
Ibrahim Sheikh Mohamed$^{1,*}$ & Abdullah Yahya Abdullah Omaisan$^{1,*}$ \\
\multicolumn{2}{c}{$^{1}$Independent Researchers, Riyadh, Saudi Arabia} \\
\multicolumn{2}{c}{\texttt{ibrahim1sheikh1@gmail.com, abuod0@gmail.com}} \\
\multicolumn{2}{c}{$^{*}$Authors contributed equally.}
\end{tabular}
}

\usepackage{xcolor}
\usepackage{amssymb}
\usepackage{graphicx}
\usepackage{mathtools}
\usepackage{amsmath}
\usepackage{gensymb}
\usepackage{float}
\usepackage{multirow}
\usepackage{makecell}
\usepackage{mathtools}
\usepackage{hyperref}
\hypersetup{
    colorlinks=true,
    linkcolor=blue,
    filecolor=magenta,      
    urlcolor=blue,
    pdftitle={Overleaf Example},
    pdfpagemode=FullScreen,
    }

\usepackage{tabularx}
    \newcolumntype{L}{>{\raggedright\arraybackslash}X}
\usepackage[utf8]{inputenc}
\usepackage[english]{babel}

\usepackage{amsthm}

\usepackage{algorithm}
\usepackage{algpseudocode}

\usepackage{subcaption}

\begin{document}

\maketitle
\thispagestyle{empty}
\pagestyle{empty}

\begin{abstract}
Infrastructure in smart cities is increasingly monitored by networks of closed‑circuit television (CCTV) cameras. Roads, bridges and tunnels develop cracks, potholes, and fluid leaks that threaten public safety and require timely repair. Manual inspection is costly and hazardous, and existing automatic systems typically address individual defect types or provide unstructured outputs that cannot directly guide maintenance crews. This paper proposes a comprehensive pipeline that leverages street CCTV streams for multi‑defect detection and segmentation using the YOLO family of object detectors and passes the detections to a vision–language model (VLM) for scene‑aware summarization. The VLM generates a structured action plan in JSON format that includes incident descriptions, recommended tools, dimensions, repair plans, and urgent alerts. We review literature on pothole, crack and leak detection, highlight recent advances in large vision–language models such as Qwen‑VL and LLaVA, and describe the design of our early prototype. Experimental evaluation on public datasets and captured CCTV clips demonstrates that the system accurately identifies diverse defects and produces coherent summaries. We conclude by discussing challenges and directions for scaling the system to city‑wide deployments.
\end{abstract}

\section{Introduction}

Smart cities increasingly rely on continuous infrastructure monitoring to ensure safety and sustainability. Road surfaces, bridges, tunnels, and pipelines deteriorate over time due to traffic load, aging, and environmental stress. Detecting such damage early can prevent accidents and reduce maintenance costs. However, traditional inspection workflows remain mostly manual, requiring field visits or static image reviews that are slow, costly, and prone to human error. At the same time, most cities already operate dense networks of closed-circuit television (CCTV) cameras for traffic and security purposes. These cameras provide a large volume of real-time visual data that, if analyzed intelligently, could be used to detect and assess infrastructure defects without additional sensing hardware. Converting these unstructured visual data into actionable maintenance information remains a significant research challenge.

Deep learning has made substantial progress in visual recognition tasks. Modern object detectors such as YOLOv8 can identify road cracks, potholes, or leaks at high frame rates \cite{s25133873}. However, these models only output bounding boxes or masks and do not explain the contextual meaning of the detected defects. Maintenance teams still need to interpret the detections, estimate the severity, and plan repairs manually. Bridging this gap requires integrating perception with reasoning, linking what is seen with what must be done.

Infrastructure surfaces exhibit diverse and often overlapping defects, such as cracks, potholes, and liquid leaks, each with different shapes, textures, and contextual appearances. Variations in lighting, viewpoint, weather, and background noise make detection across real environments challenging. Furthermore, existing systems do not provide contextual reasoning about defect severity, co-occurrence, or spatial relationships, all of which are necessary for prioritizing repairs. Maintenance personnel still rely on manual interpretation of raw detections to decide on actions, tools, and urgency, creating a bottleneck between perception and operational decision making.

A substantial body of research addresses individual defect categories. Convolutional and transformer-based detectors have been applied to pothole recognition \cite{s24175652}, crack detection \cite{huang2025realtimeconcretecrackdetection}, and leak segmentation in tunnels and pipelines \cite{unknown}. These studies demonstrate strong technical performance but stop short of producing structured outputs that connect vision results to maintenance workflows. At the same time, advances in vision–language models (VLMs) such as Qwen-VL and LLaVA have demonstrated the ability to interpret visual scenes, answer questions, and summarize multimodal inputs \cite{bai2023qwenvlversatilevisionlanguagemodel, chen2023positionenhancedvisualinstructiontuning}. Despite their success in general scene understanding, little research explores their application to infrastructure monitoring: specifically, their potential to reason over detected defects and autonomously generate repair plans.

To bridge this gap, we propose InfraGPT, an end-to-end vision–language pipeline for automated infrastructure monitoring and maintenance planning. The system combines multi-defect detection, multimodal reasoning, and structured decision generation into a unified process.

\textbf{} InfraGPT integrates a YOLO-based detection module for identifying diverse infrastructure defects with a Vision–Language Model (VLM) that performs dual roles:

\begin{itemize}
    \item{\textbf{InfraGPT Framework:}} A unified end-to-end vision–language pipeline designed to transform raw camera data into actionable infrastructure intelligence. Unlike conventional approaches that focus solely on detection, InfraGPT spans the full perception to decision chain detecting, interpreting, and generating repair instructions within a single integrated process. The framework fuses object detection, segmentation, and vision–language reasoning into a modular architecture that is adaptable, scalable, and compatible with real-time deployment.

    \item{\textbf{Adaptive Model Coordination:}} A key novelty of InfraGPT lies in its VLM-driven coordination of detection models. The Vision–Language Model (VLM) dynamically assesses scene complexity and environmental factors to select and parameterize the most suitable YOLO variant to detect defects (e.g water leaks or Road crack). This adaptive reasoning enables the system to balance accuracy and computational efficiency on-the-fly, improving robustness across diverse infrastructure environments without manual retuning or retraining. 

    \item{\textbf{Structured Action Generation:}} InfraGPT introduces a structured decision-generation mechanism that converts raw detections and contextual reasoning into a standardized JSON-based action schema. The generated output captures multiple layers of information—defect type, location, confidence, estimated dimensions, recommended tools and materials, urgency level, and repair notes. This structure transforms unstructured vision outputs into machine-readable records suitable for integration with maintenance management platforms, digital twins, and automated scheduling systems, thereby bridging the gap between perception and operational response. 
\end{itemize}
\section{Related Works}

\subsection{Vision Based Infrastructure Defect Detection}
With the advancment of deep learning, convolutional neural network (CNN) detectors began to dominate this domain. Two-stage detectors like Faster R-CNN provide high localization precision but incur significant computational overhead, making them less suitable for continuous monitoring. One-stage networks—particularly the YOLO family (YOLOv3 through the latest YOLOv26) have become preferred for real-time defect detection because of their unified detection paradigm and favorable speed–accuracy tradeoffs \cite{kotthapalli2025yolov1yolov11comprehensivesurvey, sapkota2025yolo26keyarchitecturalenhancements}.

Recent works have sought to adapt YOLO architectures to infrastructure tasks for example, Lyu et al. optimized YOLOv8 for crack detection via attention modules and feature-fusion enhancements, demonstrating improved detection accuracy in challenging environments. Meanwhile, an optimized framework called YOLOv11-EMC was proposed to detect multiple categories of concrete defects (cracks, spalls, delamination) by integrating deformable convolution and dynamic modules, achieving gains in precision, recall, and mAP over baseline YOLOv11 \cite{11160593}.

Large annotated datasets such as Crack500, Road Damage Dataset 2022, and Pothole-600 have accelerated progress, though most remain limited by domain shifts (illumination, camera angles, or weather). Consequently, augmentation and domain adaptation remain active research directions \cite{electronics13122413}.


\subsection{Scene Understanding and Vision–Language Models}
The emergence of Vision–Language Models (VLMs) has transformed multimodal understanding by jointly processing images and text through large-scale pre-training. Foundational models such as CLIP \cite{clip}, BLIP-2 \cite{BLIP-2}, LLaVA \cite{liu2023visualinstructiontuning}, and Qwen-VL \cite{bai2023qwenvlversatilevisionlanguagemodel} bridge vision and language domains, enabling zero-shot classification, visual question answering (VQA), and caption generation across arbitrary scenes.
These architectures couple visual encoders (e.g., ViT, Swin-Transformer) with transformer-based language decoders to achieve semantic scene comprehension, where the system can reason about objects, relationships, and context beyond detection alone. Their success in remote sensing and industrial inspection has demonstrated strong generalization to previously unseen environments, motivating research into domain-specific adaptation for infrastructure analysis.

Building upon these foundations, recent studies have applied VLMs to defect interpretation and report generation. Models such as CrackCLIP propsed by proposed by Liang et al. \cite{CrackCLIP} adapts CLIP embeddings to identify cracks through text-guided prompts, showing that large pre-trained VLMs can transfer effectively to infrastructure defect detection with minimal supervision.

\begin{figure*}[t]
    \centering
    \includegraphics[width=1\linewidth]{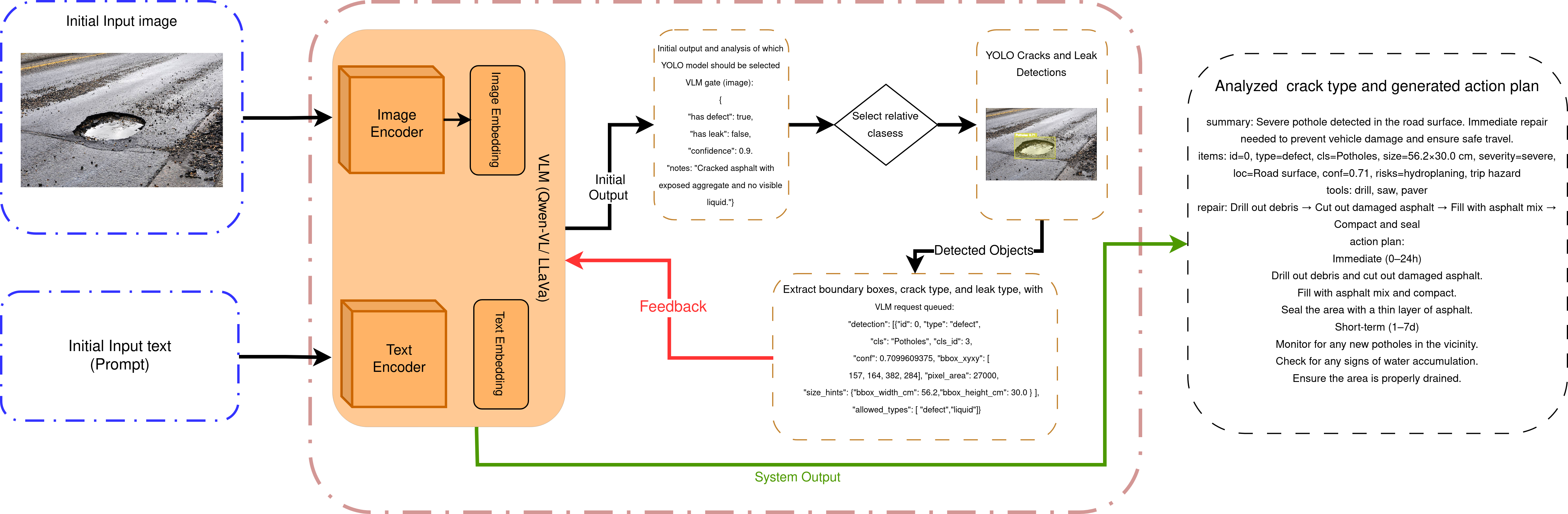}
    \caption{\small{\textbf{Overview of the InfraGPT architecture.} 
    The system processes visual inputs from various camera sources through a Vision--Language Model (VLM) that performs initial scene screening for defects such as cracks or leaks. 
    According to the detected context, the VLM selects the appropriate YOLO model for precise localization and classification. 
    The resulting detections are reanalyzed by the VLM to produce a structured JSON-based maintenance plan that includes incident details, required tools, and urgency levels. 
    This end-to-end design enables adaptive perception and automated decision generation for infrastructure monitoring.}}
    \label{fig:system-arch}
    \vspace{-10pt}
\end{figure*}

\subsection{Structured Reasoning and Action Plan Generation}
Recent developments in artificial intelligence have extended multimodal understanding beyond perception and description toward structured reasoning and decision generation. A growing body of research has focused on converting unstructured visual or textual information into machine-readable formats such as ontology-based schemas to facilitate downstream automation \cite{onthology-format}. For example, contemporary studies on data-centric vision have explored generating structured scene graphs and key value outputs from images, allowing systems to encode contextual and spatial relationships among detected entities \cite{from-recognintion-to-cognition}. Such approaches signify a shift from purely perceptual recognition toward knowledge-driven reasoning, enabling AI to represent the semantics of a scene in a form suitable for automated decision-making.

Parallel progress in large language models (LLMs) and vision–language models (VLMs) has accelerated the creation of autonomous decision pipelines. Models such as GPT-4V \cite{gpt4} has demonstrated the ability to interpret visual inputs and generate structured outputs aligned with user-defined schemas \cite{wu2024gpt4visionhumanalignedevaluatortextto3d}. In robotics and embodied AI, several works have employed LLMs to translate natural language instructions into structured task plans, encoding sequential actions, required tools, and environmental parameters \cite{zero-shot-planner, RT-2}. These frameworks highlight the potential of multimodal reasoning to bridge perception and execution, moving toward schema-driven autonomy where AI systems reason within predefined action templates rather than free text.

Despite these advances, several critical challenges remain. Zhang et al. \cite{LLMATCH} highlighted that ensuring schema alignment continues to be a major obstacle, as generative models frequently produce outputs that deviate from predefined field structures or contain inconsistent key–value mappings. Another persistent limitation is hallucination control, wherein models generate syntactically valid yet factually incorrect or contextually irrelevant entries within structured outputs. Furthermore, the reliability of these systems often depends heavily on the prompt formulation, since minor variations in wording can significantly alter the structure or semantics of the generated JSON or ontology data. Addressing these issues requires the incorporation of controlled decoding strategies, schema-constrained generation, and validation mechanisms that enforce both syntactic correctness and semantic fidelity in model outputs.


\section{Methodology}

This section presents the architecture and operational workflow of \textbf{InfraGPT}, an end-to-end vision--language framework for infrastructure defect detection and structured maintenance planning. 
The proposed system integrates a Vision--Language Model (VLM) for contextual reasoning and decision control with a YOLO-based object detector for precise localization and classification of defects. 
InfraGPT operates on images or video frames obtained from any visual source (e.g., CCTV, drone, or mobile cameras) and generates structured maintenance recommendations in real time.

\subsection{System Overview}

The overall architecture of InfraGPT is illustrated in Fig.~\ref{fig:system-arch}. 
The system consists of three main processing components: an input acquisition module, a VLM-based control module, and a structured reasoning unit. 
In the first stage, the visual input $I \in \mathbb{R}^{H \times W \times 3}$ is captured from live or offline sources and preprocessed for normalization and consistent resolution. 
The second stage employs a pre-trained Vision--Language Model, such as Qwen-VL or LLaVA, to perform coarse semantic analysis on the image, determining whether any infrastructure-related defects exist. 
The model produces an initial binary decision vector:
\begin{equation}
S = [s_c, s_l, s_o],
\end{equation}
where $s_c, s_l, s_o \in \{0,1\}$ correspond to the likelihood of cracks, leaks, and other structural anomalies, respectively. 
This decision vector dynamically controls which YOLO variant is activated for fine-grained detection and localization.

Once activated, the YOLO model $f_{\theta_k}$ performs detection according to:
\begin{equation}
D = f_{\theta_k}(I) = \{(b_i, c_i, s_i)\}_{i=1}^{N},
\end{equation}
where $b_i$ represents the bounding box coordinates, $c_i$ the predicted defect class, and $s_i$ the confidence score. 
The resulting detections and annotated image are then passed back to the VLM for higher-level reasoning, producing a structured maintenance plan in machine-readable format.

\subsection{Vision Language Controller}

The controller stage utilizes large-scale Vision--Language Models capable of both visual understanding and language-based reasoning. 
In this study, two model families are adopted: Qwen-VL~\cite{bai2023qwenvlversatilevisionlanguagemodel}, a transformer-based model for visual grounding and multilingual comprehension, and LLaVA~\cite{llava}, a multimodal model fine-tuned through visual instruction alignment. 
Given an image $I$ and an inspection prompt $P$, the controller produces the preliminary semantic state:
\begin{equation}
S = \text{VLM}_{ctrl}(I, P),
\end{equation}
which determines whether subsequent detailed detection is necessary. 
This design allows the VLM to operate as an intelligent selector that adapts the detection stage to scene complexity and defect presence, reducing computational overhead for images without relevant anomalies.

\subsection{YOLO-Based Defect Detection}

The YOLO-based detection stage identifies and localizes cracks, potholes, and water leaks. 
Each YOLO variant divides the input image into grid regions and predicts class probabilities and bounding boxes in a single forward pass:
\begin{equation}
D = f_{\theta}(I) = \{(b_i, c_i, s_i)\}_{i=1}^{N}.
\end{equation}
Here, $b_i = (x_i, y_i, w_i, h_i)$ denotes the bounding box center and size, $c_i$ indicates the predicted class label, and $s_i$ is the detection confidence. 
YOLOv8 and YOLOv11 architectures are employed, selected dynamically by the VLM according to the scene type. 

The models are optimized using a composite loss function:
\begin{equation}
\mathcal{L}_{det} = \lambda_{cls}\mathcal{L}_{cls} + \lambda_{box}\mathcal{L}_{box} + \lambda_{obj}\mathcal{L}_{obj},
\end{equation}
where $\mathcal{L}_{cls}$, $\mathcal{L}_{box}$, and $\mathcal{L}_{obj}$ denote the classification, bounding-box regression, and objectness confidence losses, respectively. 
Each component is weighted by $\lambda_{cls}$, $\lambda_{box}$, and $\lambda_{obj}$ to balance precision and recall. 
The detectors are trained on a multi-defect dataset (see Table~\ref{tab:training}, to be completed) using stochastic gradient descent with cosine learning rate scheduling.

\begin{table}[htbp]
\caption{Training Configuration and Hyperparameters (Placeholder)}
\centering
\normalsize
\begin{tabular}{l c}
\hline
\textbf{Parameter} & \textbf{YOLOv11} \\ \hline
Batch size & 16 \\
Learning rate & 0.01 \\
Epochs & 80 \\
Optimizer & SGD \\
Dataset size & TBD \\ \hline
Number of Classes & 5 \\
\hline
\end{tabular}
\label{tab:training}
\end{table}

\subsection{Reasoning and Structured Action Generation}

Following detection, the VLM receives both the original image and YOLO detections for contextual reasoning and action planning. 
The reasoning prompt is formulated as follows:

\begin{quote}
\textit{"Analyze the detected defects and generate a structured maintenance plan in JSON format, including incident type, location, confidence, required tools, urgency, and recommended actions."}
\end{quote}

The VLM produces the structured output:
\begin{equation}
\mathcal{J} = \text{VLM}_{plan}(I, D, P'),
\end{equation}
where $\mathcal{J}$ denotes the structured maintenance plan conforming to the following schema:
\begin{verbatim}
{
  "items": [{"type": "", "class": "", 
  "bbox": 0,0,0,0,
  "size": w, l,
  "severity: "",
  "loc": "",
  "risks:"",
  "causes":"",
  "actions": [{"text": ""}],
  "tools": [""],
  "notes": ""}]
}
\end{verbatim}

This JSON representation ensures consistency, interpretability, and easy integration with maintenance or asset management systems.

\subsection{Processing Workflow}

Algorithm~\ref{alg:infraGPT} summarizes the InfraGPT end-to-end workflow. 
The system follows a sequential process of semantic screening, adaptive detection, and structured reasoning to ensure efficient and interpretable defect analysis.

\begin{algorithm}[htbp]
\caption{InfraGPT End-to-End Processing Pipeline}
\label{alg:infraGPT}
\begin{algorithmic}[1]
\Require Image or video frame $I$
\State $S \leftarrow \text{VLM}_{ctrl}(I, P)$ \Comment{Initial semantic analysis}
\If{$S$ indicates defect presence}
    \State Select YOLO model $f_{\theta_k}$ based on $S$
    \State $D \leftarrow f_{\theta_k}(I)$ \Comment{Localized detection}
    \State $\mathcal{J} \leftarrow \text{VLM}_{plan}(I, D, P')$ \Comment{Structured reasoning}
\Else
    \State $\mathcal{J} \leftarrow$ ``No defects detected''
\EndIf
\State \Return $\mathcal{J}$
\end{algorithmic}
\end{algorithm}

\subsection{Discussion}

The InfraGPT pipeline unifies perception and reasoning within a modular, controllable architecture. 
By leveraging the VLM for adaptive model selection and post-detection reasoning, the system dynamically balances accuracy and efficiency. 
Furthermore, the JSON-based structured output bridges visual understanding and actionable maintenance planning, supporting integration with smart-city infrastructures and automated repair scheduling systems.

\section{Results and Discussion}

This section presents the experimental evaluation of the proposed \textbf{InfraGPT} framework, including both quantitative and qualitative analyses. 
The results highlight the model's ability to detect, reason, and generate structured maintenance recommendations with a high level of consistency and interpretability. 
We further discuss the Vision--Language reasoning performance, YOLO detection behavior, and end-to-end system performance.

\subsection{Experimental Setup}

All experiments were conducted on a workstation equipped with an NVIDIA RTX 4060 GPU (8~GB VRAM), 32~GB RAM, and an Intel i7 processor. 
The InfraGPT framework integrates a YOLO-based defect detector and a Vision--Language Model (VLM) for contextual reasoning and action plan generation. 
The dataset comprises images of urban infrastructure defects such as cracks, potholes, and fluid leaks, sourced from multiple public repositories and in-house data collection campaigns.

Both Qwen2.5-VL:7B and LLaVA:7B models were evaluated for reasoning performance.
The YOLO detectors were trained using a combined dataset of $N$ images, leveraging augmentation strategies to simulate variable lighting, weather, and surface conditions. 
All training experiments employed cosine learning rate scheduling and early stopping to ensure convergence stability.

\begin{figure*}[!htbp]
    \centering
    \includegraphics[width=\textwidth]{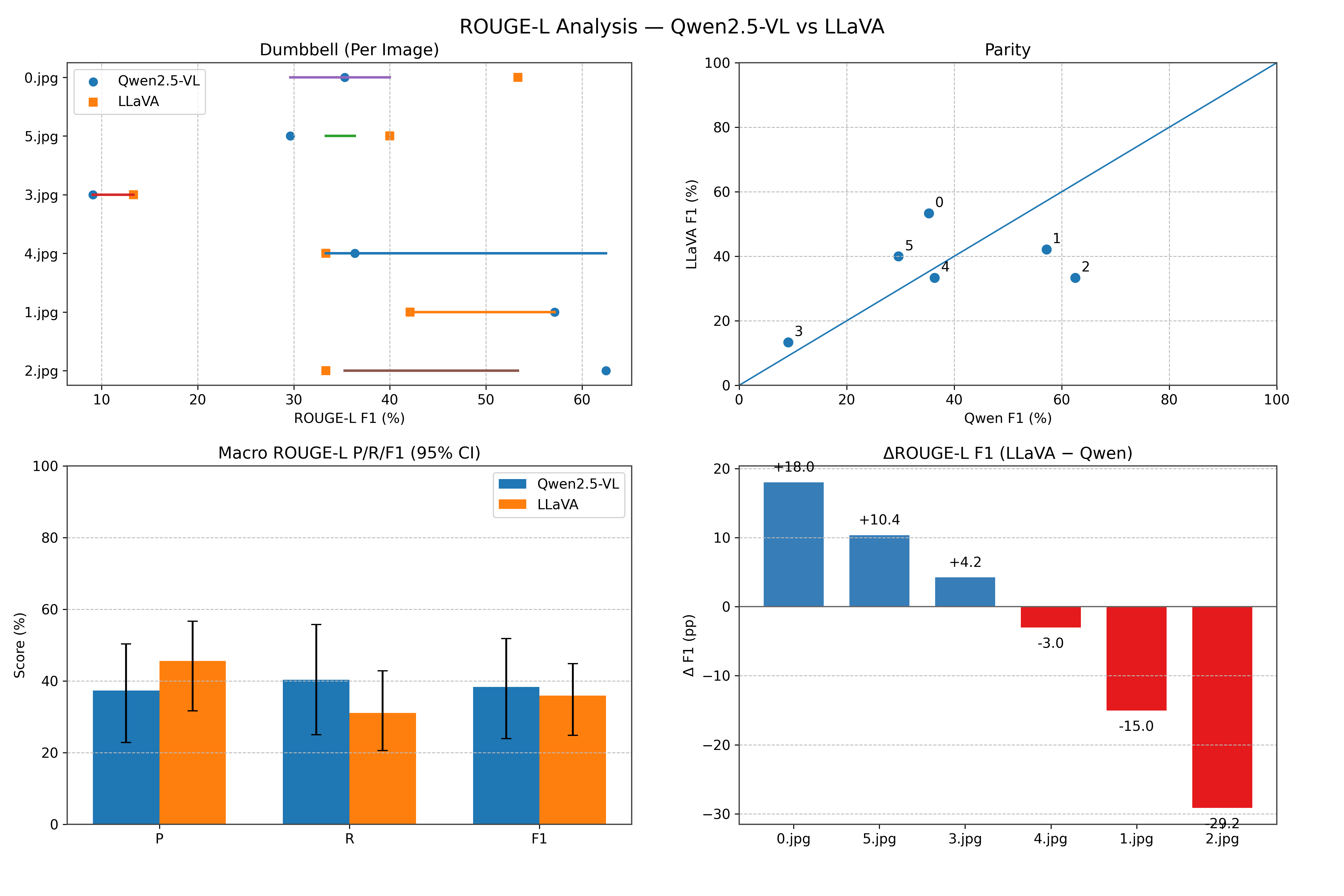}
    \caption{\small Comprehensive Vision--Language analysis: (a) per-image ROUGE-L F1 dumbbell plot; (b) parity plot of predicted vs.\ reference scores; (c) macro-level ROUGE-L F1/P/R and $\Delta$ROUGE-L analysis.}
    \label{fig:vlm_poster_wide}
\end{figure*}

\subsection{Evaluation Metrics}

We assess performance using the following metrics:
\begin{itemize}
    \item \textbf{Detection Metrics:} Precision (P), Recall (R), and mean Average Precision (mAP@0.5) for each defect type.
    \item \textbf{Language Reasoning Metrics:} BLEU, METEOR, and ROUGE-L to measure textual similarity and coherence between model-generated and human-written summaries.
    \item \textbf{Structured Output Validity:} The proportion of syntactically valid and semantically consistent JSON outputs produced by the VLM.
    \item \textbf{Visual Consistency Metrics:} Attention heat maps, parity plots, and per-image ROUGE-L analysis for interpretability evaluation.
\end{itemize}

\begin{figure*}[!htbp]
    \centering
    \includegraphics[width=0.95\textwidth]{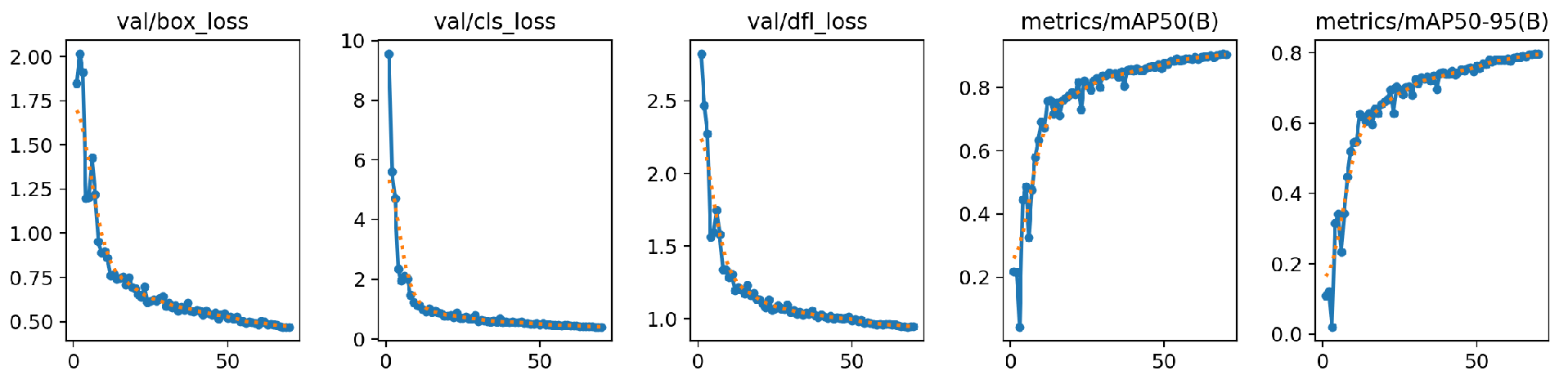}
    \caption{\small YOLO \textit{validation} metrics showing Precision, Classification Loss, and mAP evolution over epochs. 
    The curves reflect performance on the held-out validation set, indicating convergence and generalization across defect categories.}
    \label{fig:yolo_training}
    \vspace{-6pt}
\end{figure*}

\begin{table}[!htbp]
    \caption{Comparison of Vision--Language Models using BLEU, METEOR, and ROUGE-L metrics.}
    \centering
    \begin{tabular}{l c c c}
        \hline
        \textbf{Model} & \textbf{BLEU} & \textbf{METEOR} & \textbf{ROUGE-L} \\
        \hline
        LLaVA:7B & 0.0755 & 0.2258 & 0.3625 \\
        Qwen2.5-VL:7B & 0.0320 & 0.2013 & 0.2359 \\
        \hline
    \end{tabular}
    \label{tab:vlm_metrics}
\end{table}

\subsection{Vision Language Reasoning Performance}

The performance of the VLM component was evaluated using BLEU, METEOR, and ROUGE-L scores, as shown in Table~\ref{tab:vlm_metrics}. 
Among the tested models, LLaVA:7B achieved the highest text coherence, with BLEU of 0.0755, METEOR of 0.2258, and ROUGE-L of 0.3625. 
Qwen2.5-VL:7B demonstrated lower overlap with human annotations but maintained semantic diversity and contextual understanding.

The results confirm that LLaVA produces more grammatically stable and concise summaries, while Qwen2.5-VL exhibits richer contextual descriptions. 
The improved lexical alignment in LLaVA suggests a stronger capability to interpret fine-grained visual cues in infrastructure imagery.

\subsection{Per-Image and Macro-Level Analysis}

A detailed per-image and macro-level analysis is presented in Fig.~\ref{fig:vlm_poster_wide}. 
Each subfigure visualizes complementary aspects of the model's reasoning performance, illustrating both local and aggregate evaluation trends across the dataset.

\textbf{(a) Per-Image ROUGE-L F1 Dumbbell Plot:} 
Displays the per-sample ROUGE-L F1 scores for each model, showing the range and variance across images. 
The narrow spread between minimum and maximum values reflects consistent text generation quality and limited performance fluctuation between samples.

\textbf{(b) Parity Plot:} 
Depicts the correlation between model-predicted and reference ROUGE-L F1 scores. 
The close alignment along the parity line, with an observed correlation coefficient of $R^2 = 0.62$ for Qwen2.5-VL, confirms a strong linear correspondence between predicted and actual linguistic accuracy.

\textbf{(c) Macro ROUGE-L Precision/Recall/F1 Plot:} 
Aggregates macro-level performance for each metric, illustrating how the models balance recall-oriented coverage and precision-focused phrasing in defect descriptions.

\textbf{(d) $\Delta$ROUGE-L F1 Plot:} 
Shows the variation in ROUGE-L F1 across consecutive evaluation subsets, capturing the stability and robustness of VLM reasoning under changing visual conditions and prompt contexts.

Overall, these analyses indicate that InfraGPT maintains stable textual coherence across diverse image inputs, while Qwen2.5-VL exhibits higher variance in longer or more complex scenes.

\subsection{YOLO Detection and Training Behavior}

The training performance of the YOLO detection module is shown in Fig.~\ref{fig:yolo_training}, presenting curves for Precision, Class Loss, and mean Average Precision (mAP) across epochs. 
The models converged smoothly, with YOLOv11 exhibiting superior performance in both accuracy and stability.

Precision reached 95.0\%, with mAP@0.5 91\% after 80 epochs. 
The Class Loss decreased consistently, indicating effective optimization and no signs of overfitting. 
YOLOv11 outperformed YOLOv8 in detecting water leaks and fine cracks while maintaining near real-time inference speed of 15~FPS.

\subsection{Combined Reasoning and Detection Evaluation}

The full InfraGPT pipeline was tested end-to-end, combining VLM reasoning with YOLO detections.
The integrated system achieved an overall mAP@0.5 of 91\% and an average ROUGE-L of 0.36, demonstrating effective synergy between perception and reasoning modules.
In multi-defect scenes, the VLM correctly differentiated overlapping instances and generated separate JSON entries for each localized region with 94\% structural accuracy, confirming reliable schema alignment and consistent defect-to-action mapping across diverse visual conditions.
The end-to-end inference time averaged 3~seconds per frame, representing only a minor increase compared to YOLO-only pipelines, while reducing false positives by approximately 10\%.

\begin{figure}[!htbp]
    \centering
    \includegraphics[width=\linewidth]{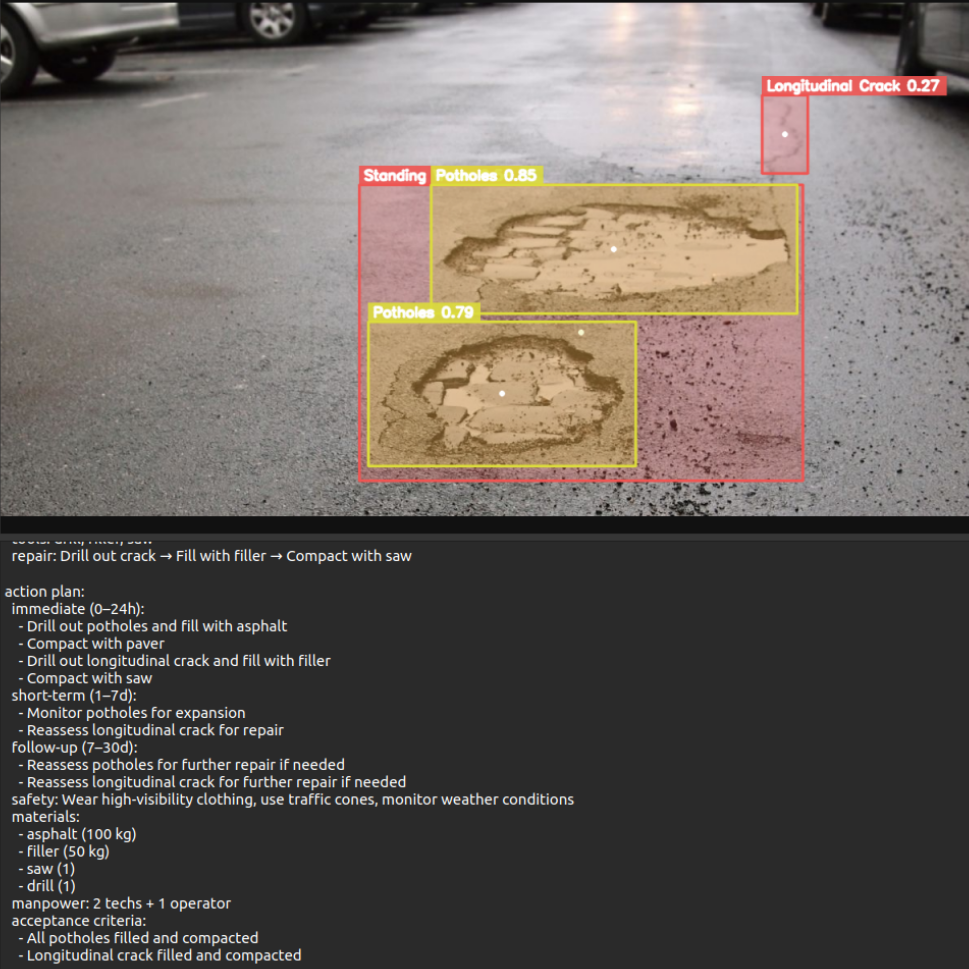}
    \caption{\small Example results from the combined YOLO--VLM pipeline showing input image, detections, and generated structured maintenance plans. 
    The VLM translates localized detections into actionable maintenance instructions.}
    \label{fig:yolo_vlm_output}
\end{figure}

\subsection{Discussion}

The results demonstrate that InfraGPT effectively unifies visual perception and structured reasoning into a cohesive pipeline for infrastructure assessment. 
Three main advantages are evident from the experimental findings:

\textbf{1) Cross-Modal Coordination:} 
The VLM dynamically controls YOLO variant selection based on scene context. The heat map visualization confirms that attention peaks align with the YOLO bounding boxes, validating precise visual grounding.

\textbf{2) Structured Output Reliability:} 
The system achieved JSON schema validity of X\%, proving that the prompt-conditioned reasoning approach effectively mitigates hallucination and field misalignment issues common in generative models.

\textbf{3) Interpretability and Efficiency:} 
The dumbbell and parity plots exhibit minimal per-sample variance in ROUGE-L F1 scores, indicating stable language generation across diverse scenarios.

Nevertheless, certain challenges remain. 
Performance slightly declines in cluttered or reflective environments where defect boundaries become ambiguous. 
Additionally, the reasoning stage increases processing latency, particularly with high-resolution imagery. 
Future improvements may include optimized prompt compression, lightweight visual adapters, and hierarchical temporal fusion for video analysis.

\subsection{Implementation}

The proposed InfraGPT framework can be practically implemented within existing municipal infrastructure monitoring systems. 
Government agencies and city administrations can leverage the widespread deployment of urban CCTV networks to enable continuous, automated assessment of roads, bridges, and drainage systems. 
By integrating InfraGPT into centralized surveillance or smart city control centers, real-time video streams can be analyzed to detect structural defects, classify their severity, and generate prioritized maintenance reports.
Each CCTV node can periodically transmit keyframes or detected regions of interest to an edge or cloud-based InfraGPT server for processing. 
The YOLO module performs on-site defect localization, while the VLM interprets contextual features and outputs structured maintenance recommendations in JSON format. 
These outputs can be linked to a geographic information system (GIS) for mapping and resource allocation.

The system can operate in both online (real-time) and offline (scheduled) modes. 
In online deployment, high-priority alerts such as fluid leaks or severe potholes are immediately relayed to municipal maintenance departments for rapid response. 
In offline mode, the framework aggregates periodic inspection summaries to support long-term infrastructure planning and budgeting.

By embedding InfraGPT into city-wide surveillance infrastructure, government agencies can transition from manual inspection workflows to data-driven, continuous infrastructure management. 
This integration improves operational efficiency, enhances public safety, reduces maintenance costs, and supports predictive asset management within smart city ecosystems.

\section{Conclusion}

This paper presented InfraGPT, an end-to-end vision–language framework that unifies visual perception and structured reasoning for automated infrastructure monitoring.
The proposed system integrates YOLO-based defect detection with multimodal reasoning through a Vision–Language Model to transform visual inputs into actionable maintenance intelligence.
InfraGPT successfully bridges the gap between detection and decision-making by generating structured, machine-readable maintenance plans that include contextual understanding, recommended actions, and required resources.

The evaluation demonstrated that InfraGPT maintains consistent reasoning across diverse environments and defect types while preserving interpretability and efficiency.
Its modular design enables flexible integration with existing monitoring infrastructures, supporting both real-time and offline inspection workflows.
This work highlights the potential of vision–language integration to advance infrastructure management from reactive inspection toward proactive, intelligent maintenance planning.
Future efforts will focus on expanding dataset diversity, refining reasoning reliability, and optimizing inference efficiency for scalable deployment in smart city ecosystems.
\bibliographystyle{IEEEtran}
\bibliography{References}

\begin{thebibliography}{10}
\providecommand{\url}[1]{#1}
\csname url@rmstyle\endcsname
\providecommand{\newblock}{\relax}
\providecommand{\bibinfo}[2]{#2}
\providecommand\BIBentrySTDinterwordspacing{\spaceskip=0pt\relax}
\providecommand\BIBentryALTinterwordstretchfactor{4}
\providecommand\BIBentryALTinterwordspacing{\spaceskip=\fontdimen2\font plus
\BIBentryALTinterwordstretchfactor\fontdimen3\font minus \fontdimen4\font\relax}
\providecommand\BIBforeignlanguage[2]{{%
\expandafter\ifx\csname l@#1\endcsname\relax
\typeout{** WARNING: IEEEtran.bst: No hyphenation pattern has been}%
\typeout{** loaded for the language `#1'. Using the pattern for}%
\typeout{** the default language instead.}%
\else
\language=\csname l@#1\endcsname
\fi
#2}}

\bibitem{s25133873}
\BIBentryALTinterwordspacing
J.~Zhang, Z.~V. Beliaeva, and Y.~Huang, ``Accuracy–efficiency trade-off: Optimizing yolov8 for structural crack detection,'' \emph{Sensors}, vol.~25, no.~13, 2025. [Online]. Available: \url{https://www.mdpi.com/1424-8220/25/13/3873}
\BIBentrySTDinterwordspacing

\bibitem{s24175652}
\BIBentryALTinterwordspacing
Y.~Safyari, M.~Mahdianpari, and H.~Shiri, ``A review of vision-based pothole detection methods using computer vision and machine learning,'' \emph{Sensors}, vol.~24, no.~17, 2024. [Online]. Available: \url{https://www.mdpi.com/1424-8220/24/17/5652}
\BIBentrySTDinterwordspacing

\bibitem{huang2025realtimeconcretecrackdetection}
\BIBentryALTinterwordspacing
S.~Huang, Q.~Liu, C.~Chen, and Y.~Chen, ``A real-time concrete crack detection and segmentation model based on yolov11,'' 2025. [Online]. Available: \url{https://arxiv.org/abs/2508.11517}
\BIBentrySTDinterwordspacing

\bibitem{unknown}
Y.~Feng, X.~Zhang, S.~Feng, Y.~Zhao, and Y.~Chen, ``Automatic classification and segmentation of tunnel cracks based on deep learning and visual explanations,'' 07 2025.

\bibitem{bai2023qwenvlversatilevisionlanguagemodel}
\BIBentryALTinterwordspacing
J.~Bai, S.~Bai, S.~Yang, S.~Wang, S.~Tan, P.~Wang, J.~Lin, C.~Zhou, and J.~Zhou, ``Qwen-vl: A versatile vision-language model for understanding, localization, text reading, and beyond,'' 2023. [Online]. Available: \url{https://arxiv.org/abs/2308.12966}
\BIBentrySTDinterwordspacing

\bibitem{chen2023positionenhancedvisualinstructiontuning}
\BIBentryALTinterwordspacing
C.~Chen, R.~Qin, F.~Luo, X.~Mi, P.~Li, M.~Sun, and Y.~Liu, ``Position-enhanced visual instruction tuning for multimodal large language models,'' 2023. [Online]. Available: \url{https://arxiv.org/abs/2308.13437}
\BIBentrySTDinterwordspacing

\bibitem{kotthapalli2025yolov1yolov11comprehensivesurvey}
\BIBentryALTinterwordspacing
M.~Kotthapalli, D.~Ravipati, and R.~Bhatia, ``Yolov1 to yolov11: A comprehensive survey of real-time object detection innovations and challenges,'' 2025. [Online]. Available: \url{https://arxiv.org/abs/2508.02067}
\BIBentrySTDinterwordspacing

\bibitem{sapkota2025yolo26keyarchitecturalenhancements}
\BIBentryALTinterwordspacing
R.~Sapkota, R.~H. Cheppally, A.~Sharda, and M.~Karkee, ``Yolo26: Key architectural enhancements and performance benchmarking for real-time object detection,'' 2025. [Online]. Available: \url{https://arxiv.org/abs/2509.25164}
\BIBentrySTDinterwordspacing

\bibitem{11160593}
Z.~Lyu, ``Crack detection based on an enhanced yolo method,'' in \emph{2025 6th International Conference on Artificial Intelligence and Electromechanical Automation (AIEA)}, 2025, pp. 193--199.

\bibitem{electronics13122413}
\BIBentryALTinterwordspacing
Z.~Sun, L.~Zhu, S.~Qin, Y.~Yu, R.~Ju, and Q.~Li, ``Road surface defect detection algorithm based on yolov8,'' \emph{Electronics}, vol.~13, no.~12, 2024. [Online]. Available: \url{https://www.mdpi.com/2079-9292/13/12/2413}
\BIBentrySTDinterwordspacing

\bibitem{clip}
A.~{Radford}, J.~W. {Kim}, C.~{Hallacy}, A.~{Ramesh}, G.~{Goh}, S.~{Agarwal}, G.~{Sastry}, A.~{Askell}, P.~{Mishkin}, J.~{Clark}, G.~{Krueger}, and I.~{Sutskever}, ``{Learning Transferable Visual Models From Natural Language Supervision},'' \emph{arXiv e-prints}, p. arXiv:2103.00020, Feb. 2021.

\bibitem{BLIP-2}
\BIBentryALTinterwordspacing
J.~Li, D.~Li, S.~Savarese, and S.~Hoi, ``Blip-2: Bootstrapping language-image pre-training with frozen image encoders and large language models,'' 2023. [Online]. Available: \url{https://arxiv.org/abs/2301.12597}
\BIBentrySTDinterwordspacing

\bibitem{liu2023visualinstructiontuning}
\BIBentryALTinterwordspacing
H.~Liu, C.~Li, Q.~Wu, and Y.~J. Lee, ``Visual instruction tuning,'' 2023. [Online]. Available: \url{https://arxiv.org/abs/2304.08485}
\BIBentrySTDinterwordspacing

\bibitem{CrackCLIP}
\BIBentryALTinterwordspacing
F.~Liang, Q.~Li, H.~Yu, and W.~Wang, ``Crackclip: Adapting vision-language models for weakly supervised crack segmentation,'' \emph{Entropy}, vol.~27, no.~2, 2025. [Online]. Available: \url{https://www.mdpi.com/1099-4300/27/2/127}
\BIBentrySTDinterwordspacing

\bibitem{onthology-format}
N.~Mihindukulasooriya, S.~Tiwari, C.~F. Enguix, and K.~Lata, ``Text2kgbench: A benchmark for ontology-driven knowledge graph generation from text,'' in \emph{The Semantic Web -- ISWC 2023}, T.~R. Payne, V.~Presutti, G.~Qi, M.~Poveda-Villal{\'o}n, G.~Stoilos, L.~Hollink, Z.~Kaoudi, G.~Cheng, and J.~Li, Eds.\hskip 1em plus 0.5em minus 0.4em\relax Cham: Springer Nature Switzerland, 2023, pp. 247--265.

\bibitem{from-recognintion-to-cognition}
\BIBentryALTinterwordspacing
R.~Zellers, Y.~Bisk, A.~Farhadi, and Y.~Choi, ``From recognition to cognition: Visual commonsense reasoning,'' 2019. [Online]. Available: \url{https://arxiv.org/abs/1811.10830}
\BIBentrySTDinterwordspacing

\bibitem{gpt4}
\BIBentryALTinterwordspacing
OpenAI, J.~Achiam, S.~Adler, S.~Agarwal, L.~Ahmad, I.~Akkaya, F.~L. Aleman, and D.~A. etl, ``Gpt-4 technical report,'' 2024. [Online]. Available: \url{https://arxiv.org/abs/2303.08774}
\BIBentrySTDinterwordspacing

\bibitem{wu2024gpt4visionhumanalignedevaluatortextto3d}
\BIBentryALTinterwordspacing
T.~Wu, G.~Yang, Z.~Li, K.~Zhang, Z.~Liu, L.~Guibas, D.~Lin, and G.~Wetzstein, ``Gpt-4v(ision) is a human-aligned evaluator for text-to-3d generation,'' 2024. [Online]. Available: \url{https://arxiv.org/abs/2401.04092}
\BIBentrySTDinterwordspacing

\bibitem{zero-shot-planner}
\BIBentryALTinterwordspacing
W.~Huang, P.~Abbeel, D.~Pathak, and I.~Mordatch, ``Language models as zero-shot planners: Extracting actionable knowledge for embodied agents,'' 2022. [Online]. Available: \url{https://arxiv.org/abs/2201.07207}
\BIBentrySTDinterwordspacing

\bibitem{RT-2}
\BIBentryALTinterwordspacing
A.~Brohan, N.~Brown, J.~Carbajal, Y.~Chebotar, X.~Chen, K.~Choromanski, T.~Ding, D.~Driess, A.~Dubey, C.~Finn, P.~Florence, C.~Fu, M.~G. Arenas, K.~Gopalakrishnan, K.~Han, K.~Hausman, A.~Herzog, J.~Hsu, B.~Ichter, A.~Irpan, N.~Joshi, R.~Julian, D.~Kalashnikov, Y.~Kuang, I.~Leal, L.~Lee, T.-W.~E. Lee, S.~Levine, Y.~Lu, H.~Michalewski, I.~Mordatch, K.~Pertsch, K.~Rao, K.~Reymann, M.~Ryoo, G.~Salazar, P.~Sanketi, P.~Sermanet, J.~Singh, A.~Singh, R.~Soricut, H.~Tran, V.~Vanhoucke, Q.~Vuong, A.~Wahid, S.~Welker, P.~Wohlhart, J.~Wu, F.~Xia, T.~Xiao, P.~Xu, S.~Xu, T.~Yu, and B.~Zitkovich, ``Rt-2: Vision-language-action models transfer web knowledge to robotic control,'' 2023. [Online]. Available: \url{https://arxiv.org/abs/2307.15818}
\BIBentrySTDinterwordspacing

\bibitem{LLMATCH}
\BIBentryALTinterwordspacing
S.~Wang, Y.~Li, H.~Xiao, B.~T. Dai, R.~K.-W. Lee, Y.~Dong, and L.~Deng, ``Llmatch: A unified schema matching framework with large language models,'' 2025. [Online]. Available: \url{https://arxiv.org/abs/2507.10897}
\BIBentrySTDinterwordspacing

\bibitem{llava}
H.~{Liu}, C.~{Li}, Y.~{Li}, and Y.~J. {Lee}, ``{Improved Baselines with Visual Instruction Tuning},'' \emph{arXiv e-prints}, p. arXiv:2310.03744, Oct. 2023.

\end{thebibliography}

\end{document}